\documentclass[letterpaper, 10 pt, conference]{ieeeconf}  % Comment this line out if you need a4paper

\IEEEoverridecommandlockouts                              % This command is only needed if 
\usepackage{times}
\usepackage{epsfig}
\usepackage{graphicx}
\usepackage{amsmath}
\usepackage{amssymb}
\usepackage{subcaption}
\usepackage{algorithmicx}
\usepackage{algpseudocode}
\usepackage{leftidx}
\usepackage{booktabs}
\usepackage[percent]{overpic}
\usepackage{color}
\usepackage[dvipsnames]{xcolor}
\usepackage{tikz}
\usetikzlibrary{positioning}
\graphicspath{{./images/}}

\title{\LARGE \bf
Keyframe-Based Visual-Inertial Online SLAM with Relocalization
}

\author{Anton Kasyanov, Francis Engelmann, J\"org St\"uckler and Bastian Leibe% <-this % stops a space
\thanks{This  work  was  funded  by  ERC  Starting  Grant  project  CV-SUPER  (ERC-2012-StG-307432).}% <-this % stops a space
\thanks{All authors are with the Computer Vision Group, Visual Computing Institute, RWTH Aachen University,
		52074 Aachen, Germany
        {\tt\small anton.kasyanov@rwth-aachen.de, \{ engelmann, stueckler, leibe \}@vision.rwth-aachen.de}%
}
}
\begin{document}

\maketitle
\thispagestyle{empty}
\pagestyle{empty}

%%%%%%%%%%%%%%%%%%%%%%%%%%%%%%%%%%%%%%%%%%%%%%%%%%%%%%%%%%%%%%%%%%%%%%%%%%%%%%%%
\begin{abstract}
Complementing images with inertial measurements has become one of the most popular approaches to achieve highly accurate and robust real-time camera pose tracking.
In this paper, we present a keyframe-based approach to visual-inertial simultaneous localization and mapping (SLAM) for monocular and stereo cameras.
Our visual-inertial SLAM system is based on a real-time capable visual-inertial odometry method that provides locally consistent trajectory and map estimates.
We achieve global consistency in the estimate through online loop-closing and non-linear optimization.
Furthermore, our system supports relocalization in a map that has been previously obtained and allows for continued SLAM operation.
We evaluate our approach in terms of accuracy, relocalization capability and run-time efficiency on public indoor benchmark datasets and on newly recorded outdoor sequences.
We demonstrate state-of-the-art performance of our system compared to a visual-inertial odometry method and baseline visual SLAM approaches in recovering the trajectory of the camera.

\end{abstract}

%%%%%%%%%%%%%%%%%%%%%%%%%%%%%%%%%%%%%%%%%%%%%%%%%%%%%%%%%%%%%%%%%%%%%%%%%%%%%%%%
\section{Introduction}

Visual-inertial camera pose tracking (i.e. odometry) has recently attracted significant attention in the computer vision, augmented reality, and robotics communities.
The two sensor types, camera and inertial measurement unit (IMU), complement each other well in a joint optimization framework.
Visual sensors provide rich information for robust visual tracking and allow for referencing the camera trajectory towards previously visited parts of the environment.
Using vision, the full 3D rotation and translation of the camera can be observed if the environment provides sufficient features.
IMUs, on the other hand, measure accelerations and rotational velocities at high frame-rates.
This allows for overcoming degenerate visual cases such as pointing the camera at a textureless wall.
IMUs also observe acceleration due to gravity and, hence, allow for an absolute horizontal reference in the environment.

\begin{figure}[t]
	\vspace{3mm}
    \includegraphics[width=0.99\linewidth]{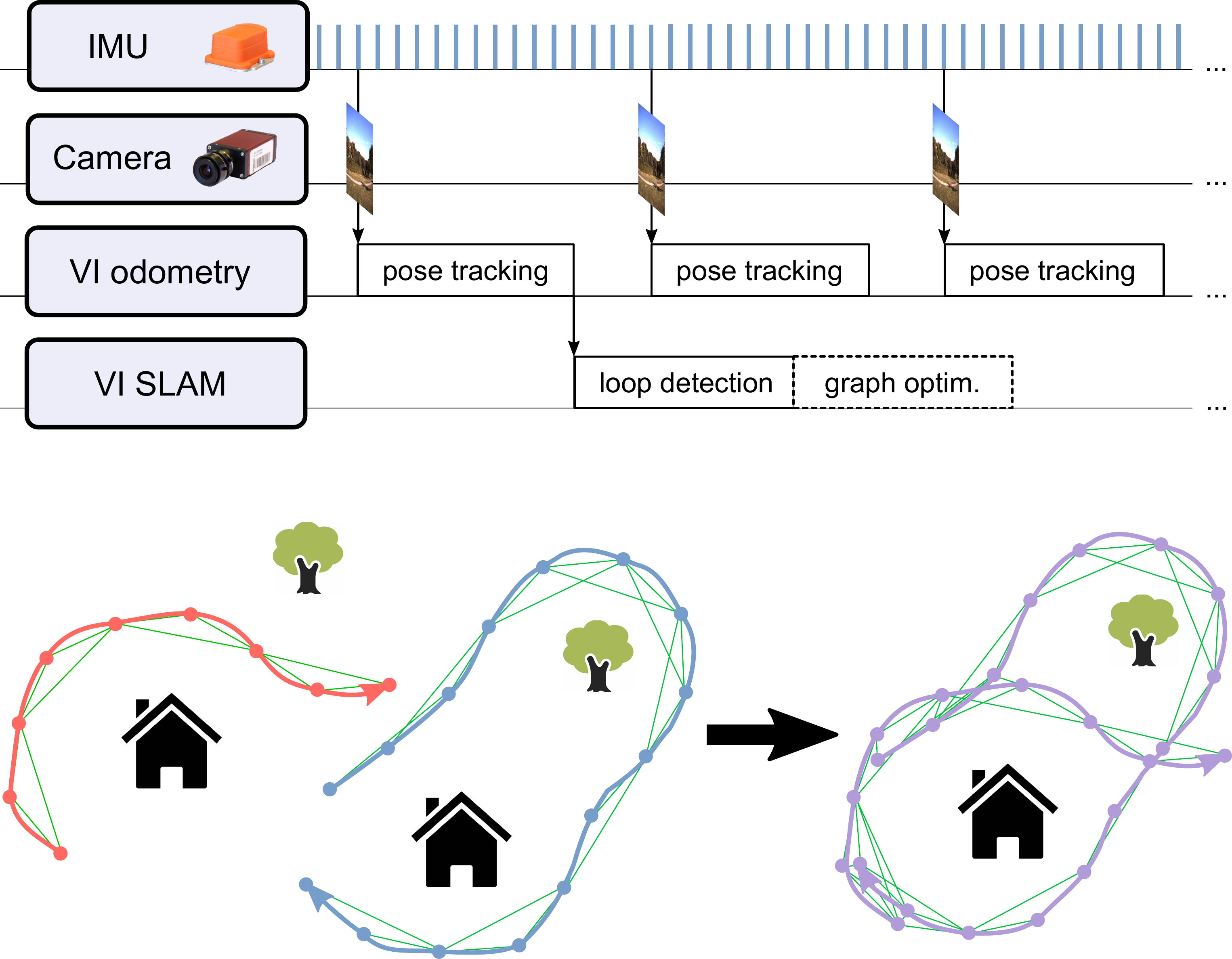}
    \vspace{3.25mm}
    \caption{Our visual-inertial SLAM system performs visual-inertial camera pose tracking and keyframe-based pose graph optimization in parallel. Top: Overview of the individual input and processing steps of our SLAM approach and their timing. Bottom: Our system allows for relocalization in a previous keyframe map and continued SLAM operation. Keyframes (dots) along  the previous (red) and new trajectory (blue) are continously merged into a combined SLAM map (purple) through loop closure constraints (green lines).}
    \label{fig:teaser}
\end{figure}

Visual-inertial odometry methods, however, are prone to drift since they typically only optimize a trajectory over a local set of frames.
In many applications such as environment mapping and autonomous robot navigation, a globally consistent trajectory and map estimate is required.
Furthermore, the system should be capable of relocalizing in a previously built map and continuing SLAM in previously unknown parts of the environment.
In this paper, we propose a real-time capable visual-inertial SLAM system that detects the revisit of locations (loop closing) using online image retrieval and optimizes the trajectory for global consistency.
We propose means to relocalize in a previously built map and to continue SLAM after relocalization (see Fig.~\ref{fig:teaser}).
Our system builds on state-of-the-art techniques to form a new baseline for visual-inertial SLAM systems.

In our experiments, we evaluate the accuracy and run-time of our SLAM approach on the EuRoC MAV benchmark dataset~\cite{burri2016_eurocmav}
and additional challenging home-made outdoor sequences.	
Using these datasets, we also demonstrate the capability of our system to relocalize and continue SLAM afterwards.

In summary, we propose a real-time capable visual-inertial SLAM system with the following properties:
(1) We achieve accurate and globally consistent visual-inertial SLAM through state-of-the-art local tracking, pose graph optimization and loop closure detection.
(2) Our SLAM approach is capable of relocalization and continued SLAM in a previously built map.

\section{Related Work}
Over the last decades, tremendous progress has been achieved in the development of visual localization and SLAM methods.
Many of the current state-of-the-art SLAM systems for monocular or stereo cameras use keyframes in order to locally track the camera motion towards a reference frame.
They optimize the camera trajectory for global consistency based on relative spatial constraints in a subsequent SLAM graph optimization layer.
Two prominent recent examples of such methods are ORB-SLAM~\cite{murartal2015_orbslam} and LSD-SLAM~\cite{engel2014_lsdslam}.

Using inertial measurements to aid visual camera tracking has been studied extensively in recent years.
Most of those approaches, however, are visual-inertial odometry methods, which perform tracking with local consistency using filtering or local window optimization approaches---inevitably accumulating drift in the camera trajectory and map estimate.
Current state-of-the-art techniques for visual-inertial odometry tightly couple visual and inertial information for state estimation~\cite{mourikis2013_msckf,leutenegger2014keyframe,usenko2016_dvio,forster2015_imupreint}.
The multi-state constraint Kalman Filter~\cite{mourikis2013_msckf} uses interest point measurements in a filtering approach to estimate camera pose, calibration and IMU biases.
The keyframe-based approach by Leutenegger et al.~\cite{leutenegger2014keyframe} optimizes camera poses and IMU biases in a local SLAM window over recent frames and keyframes.
The method tracks and reconstructs the 3D location of interest points in the images.
It marginalizes old frames in order to keep a consistent estimate of the state uncertainty.
A similar optimization approach has been proposed by Usenko et al.~\cite{usenko2016_dvio}.
They replace indirect interest-point measurements with direct image alignment in the local SLAM optimization function.
Visual-inertial SLAM approaches that provide a globally consistent map and trajectory estimate are of recent interest of the robotics and computer vision communities.
In contrast to our visual-inertial SLAM approach, the methods in~\cite{concha2016_dvi_mapping,omari2015_densevinav,ma2016_largescalevislam} do not incorporate loop closures and full SLAM optimization.
Furthermore, our approach seamlessly integrates global optimization with relocalization in an existing map.

Visual localization, e.g. on mobile phones, has typically been approached by mapping the environment in an offline process and localizing within the prebuilt map afterwards (e.g.,~\cite{lim2012_imagebasedloc,ventura2012_imagebasedloc,middelberg2014_imagebasedloc,lynen2015_viloc}).
A recent visual-inertial localization approach that runs efficiently on mobile devices has recently been proposed by Lynen et al.~\cite{lynen2015_viloc}.
This approach uses sophisticated map compression and image matching techniques to localize the camera in a prebuilt interest-point based map.
However, in contrast to our visual-inertial SLAM approach, these methods do not allow for continued global mapping of the environment using SLAM.

\section{System Overview}

Our visual-inertial SLAM approach uses visual-inertial odometry to track the 6-DoF camera motion real-time.
The visual-inertial odometry method only considers a local set of recent frames and keyframes for optimization.
While it marginalizes out the other frames along the trajectory, it still is prone to drift on the long run.
Loop closure detection and pose graph optimization is performed in a parallel thread that establishes global consistency of the trajectory estimate. % (see Fig.~\ref{fig:systemoverview}).
This thread does not need to process the frames in real-time and can possibly lag several frames behind.
In an online SLAM system, we correct the visual-inertial odometry estimate to align with the pose of the most recent keyframe that is optimized in the SLAM layer.

In order to reduce drift, the SLAM layer detects loop closures, i.e. overlapping keyframes that have not been aligned within the optimization window of the visual-inertial odometry and may match distinct landmarks in the visual odometry.
To this end, we use an image retrieval method that finds a best match for the most recent keyframe in the SLAM layer with the past keyframes.
Subsequent pose estimation between the matched keyframes using RANSAC also geometrically verifies the matching.

For relocalization in a previously built map, we use the same image retrieval and geometric alignment/verification approach as for loop-closing.
The method requires several consistent subsequent matchings before accepting the relocalization hypothesis.
We seamlessly continue the keyframe mapping using our SLAM layer mechanisms by including the new keyframes and their spatial constraints with the old map into the SLAM pose graph.

%\begin{figure}[t]
    %\includegraphics[width=\linewidth]{pipeline.png}
    %\caption{Overview of the individual input and processing steps of our SLAM approach and their timing.}
    %\label{fig:systemoverview}
%\end{figure}

\section{Keyframe-based Visual-Inertial Online SLAM with Relocalization}

\subsection{Keyframe-based Visual-Inertial Odometry}

We follow the approach by Leutenegger et al.~\cite{leutenegger2014keyframe} for visual-inertial~(VI) odometry.
In this approach, VI odometry is formulated as a non-linear optimization problem that tightly couples visual and inertial measurements:
The method extracts and matches BRISK interest points~\cite{leutenegger2011_brisk} between images.
It optimizes their 3D position as landmarks jointly with the camera poses of the images and the IMU bias parameters.
The approach uses a small set of recent frames and keyframes to make optimization feasible in real-time.
It marginalizes over older frames and keyframes to keep a consistent prior and uncertainty of the estimate.
Despite its local optimization window, the method achieves high accuracy in camera pose tracking by approximately relinearizing the marginalized terms.
Still, the method accumulates drift and does not support the detection and closing of trajectory loops.
We will handle this in an additional SLAM layer that we implemented on top of the visual-inertial odometry.
The original approach supports one or multiple cameras, whereas in our work, we focus on monocular and stereo cameras.

Formally, the method estimates the body pose, velocity and IMU bias state~$\mathbf{x}_B^k$ at frames~$k$ jointly with the position of landmarks $\mathbf{x}_L$ that are currently seen within the optimization window.
The body state variables~$\mathbf{x}_M^k = \left[ \leftidx{_{W}}{\boldsymbol{\xi}_B^T}, {}_{B}\mathbf{v}^T, \mathbf{b}_g^T, \mathbf{b}_a^T \right]^T$ comprise 6-DoF body pose~$\leftidx{_{W}}{\boldsymbol{\xi}_B}$ in the world frame (represented by 3D position and quaternion), velocity~${}_{B}\mathbf{v}$ in the body frame, the gyroscope biases $\mathbf{b}_g$ and the accelerometer biases $\mathbf{b}_a$.

Leutenegger et al.~\cite{leutenegger2014keyframe} formulate visual-inertial odometry as joint non-linear optimization of all body and landmark states~$\mathbf{x}$ using the cost function
\begin{multline}\label{eq:vio_objective}
    E_{O}(\mathbf{x}) =
        \underbrace{
            \sum_{k=1}^K \sum_{c=1}^C \sum_{l \in \mathcal{L}(c,k)}
            \mathbf{e}_V^{c,k,l}(\mathbf{x})^\top \mathbf{\Omega}_V^{c,k,l} \mathbf{e}_V^{c,k,l}(\mathbf{x})
        }_{\text{visual residuals}}\\
         + \underbrace{
            \sum_{k=1}^{K-1}\mathbf{e}_I^{k}(\mathbf{x})^\top \mathbf{\Omega}_I^k \mathbf{e}_I^k(\mathbf{x})
        }_{\text{inertial residuals}}
\end{multline}
In this formulation,~$\mathbf{e}_V$ and~$\mathbf{e}_I$ are the visual and the IMU residual terms for each frame~$k$, respectively.
The visual residuals measure the reprojection error of each visible landmark~$l \in \mathcal{L}(c,k)$ in each camera~$c$ and frame~$f$.
The IMU residuals on the other hand quantify the misalignment of the estimated camera motion with the propagated IMU measurements between frames.
The residuals are weighted by their inverse covariance through the information matrices~$\mathbf{\Omega}_V^{c,k,l}$ and~$\mathbf{\Omega}_I^k$.
The information matrices are computed as described in~\cite{leutenegger2014keyframe}.
This non-linear least squares objective can be efficiently optimized using Gauss-Newton methods.

The method distinguishes two kinds of frames: recent (intermediate) frames and keyframes.
A keyframe is created from the current frame, if the hull of the projected landmarks covers less than half of the image.
Solving the full SLAM problem including all frames, keyframes and visible landmarks in the optimization window is not feasible.
Instead, only a small amount of most recent frames and keyframes is kept in the optimization window.
The remaining frames and landmarks are marginalized out from the overall objective function in Eq.~\eqref{eq:vio_objective} which can be efficiently performed using the Schur complement.
Since state variables in the active optimization window are involved in the linearization for the marginalization, the linearization point of the marginalized states is continuously updated during the optimization.

\subsection{Keyframe-based Visual-Inertial SLAM}

Due to the optimization of a local set of frames and keyframe states, the visual-inertial odometry approach is prone to drift and cannot perform loop-closures to reduce the drift.
Hence, we turn the visual-inertial odometry method into a full SLAM method using a second layer of global pose graph optimization which is executed in parallel.

In our SLAM approach, we keep track of visited locations and detect if the same location is revisited using image retrieval techniques.
As soon as a keyframe leaves the local optimization window of the visual-inertial odometry, it is linked in the SLAM pose graph with its predecessor keyframe using the tracked pose estimate.
Additionally, we search for similar images in the database of already included keyframes in the SLAM map.
To this end, we use a state-of-the-art incremental image retrieval method, i.e. DBoW2~\cite{galvez2012_dbow2}.
We estimate and verify an accurate relative pose between the keyframes using 2D-to-3D RANSAC alignment of the landmarks in the keyframes and eventually add the spatial constraint to the SLAM graph.
By this, also the closing of longer loops can be detected and accumulated drift by the VI odometry is corrected.

We now explain in detail the pose graph optimization and the loop closure detection.

\subsubsection{Pose Graph Optimization}
\label{graph_optimization}

We distinguish two types of relative spatial constraints between the keyframe poses:
\begin{itemize}
    \item Relative pose~$\leftidx{_{k+1}}{\boldsymbol{\xi}^\mathit{seq}_{k}}$ between subsequent keyframes $k$, $k+1$ that is determined by the VI odometry.
    \item Relative pose~$\leftidx{_{k'}}{\boldsymbol{\xi}^\mathit{cls}_{k}}$ between keyframes $k$, $k'$ that were detected in a loop closure. The relative transformation between the two is determined through RANSAC 2D-to-3D matching.
\end{itemize}

We optimize the following objective function on our SLAM layer
\begin{multline}
E_{S}(\boldsymbol{\xi}) =
    \underbrace{
        \sum_{k=1}^{K-1} \mathbf{e}_{\mathit{seq}}^{k,k+1}(\boldsymbol{\xi})^\top \mathbf{\Omega}_{\mathit{seq}}^{k,k+1} \mathbf{e}_{\mathit{seq}}^{k,k+1}(\boldsymbol{\xi}) }_{\text{visual-inertial odometry residuals}}\\
    + \underbrace{
        \sum_{\left(k,k'\right) \in \mathcal{C}} \mathbf{e}_{\mathit{cls}}^{k,k'}(\boldsymbol{\xi})^\top \mathbf{\Omega}_{\mathit{cls}}^{k,k'} \mathbf{e}_{\mathit{cls}}^{k,k'}(\boldsymbol{\xi})
    }_{\text{loop closure residuals}}
\end{multline}
with
\begin{align}
	\mathbf{e}_{\mathit{seq}}^{k,k+1} &:= \leftidx{_{k+1}}{\boldsymbol{\xi}^\mathit{seq}_{k}} \ominus \left( \boldsymbol{\xi}_{k+1} \ominus   \boldsymbol{\xi}_{k} \right),\\
	\mathbf{e}_{\mathit{cls}}^{k,k'} &:= \leftidx{_{k'}}{\boldsymbol{\xi}^\mathit{cls}_{k}} \ominus \left( \boldsymbol{\xi}_{k'} \ominus \boldsymbol{\xi}_{k} \right),
\end{align}
where~$\boldsymbol{\xi}$ are the estimated poses~$\boldsymbol{\xi}_k$ of the keyframes in the SLAM graph,~$\mathcal{C}$ is the set of loop-closure constraints, and~$\mathbf{e}_{\mathit{seq}}^{k,k+1}$ and~$\mathbf{e}_{\mathit{cls}}^{k,k'}$ are the residuals for the relative pose constraints originating from sequential pose tracking and loop closing, respectively.
The operator~$\ominus$ determines the relative pose so that
\begin{align}
	\boldsymbol{\xi}_1 \ominus \boldsymbol{\xi}_2 &:= \boldsymbol{\xi}\left( \mathbf{T}\left(\boldsymbol{\xi}_1\right)^{-1} \mathbf{T}\left(\boldsymbol{\xi}_2\right) \right),
\end{align}
where~$\boldsymbol{\xi}\left( \mathbf{T} \right)$ and~$\mathbf{T}\left( \boldsymbol{\xi} \right)$ convert between pose parametrization~$\boldsymbol{\xi}$ and $\mathbf{T} \in \mathbf{SE}(3)$.
The information matrices $\mathbf{\Omega}_{\mathit{seq}}^{k,k+1}$ and $\mathbf{\Omega}_{\mathit{cls}}^{k,k'}$ model the inverse covariance of the spatial constraints.
They are approximated by isotropic covariances whose variance scales with the keyframe overlap of the frames or is constant for the loop closure constraints.
This non-linear least squares problem is optimized using the Levenberg-Marquardt method within the gtsam3 framework~\cite{dellaert2006_sqrtsam} which employs local pose parametrization using the Lie algebra of SE(3).

\subsubsection{Loop Closure Detection}
\label{sec:loopclosuredetection}

We detect loop closures in a two-stage approach:
\begin{itemize}
    \item We use DBoW2~\cite{galvez2012_dbow2} to retrieve similar images for a query keyframe from the set of keyframes in the SLAM graph.
    \item We determine the relative pose between the matched keyframes from 2D-to-3D correspondences of interest points to landmarks and using RANSAC~\cite{kneip2014_opengv}. This step also geometrically verifies the image matching found through DBoW2.
\end{itemize}

We focus our algorithm on loop closures that match keyframes which do not share common landmarks in the VI odometry.
This prevents the method to additionally match subsequent keyframes, while this information is already included directly from the VI odometry estimate.

The SLAM layer provides a pose in the global reference frame for the most recent keyframe in the map.
We use this pose to correct the real-time VI odometry estimate of the body pose.

\subsection{Relocalization and Continued SLAM}

%\begin{figure}[t]
  %\includegraphics[width=0.55\linewidth]{reloc1.pdf}\hfill
  %\includegraphics[width=0.25\linewidth]{reloc2.pdf}
  %\caption{Relocalization. Left: previous and new SLAM maps (trajectories) before relocalization. Right: combined SLAM map after relocalization.}
  %\label{fig:reloc_example}
%\end{figure}

\begin{figure}[t]
   \includegraphics[width=0.99\linewidth]{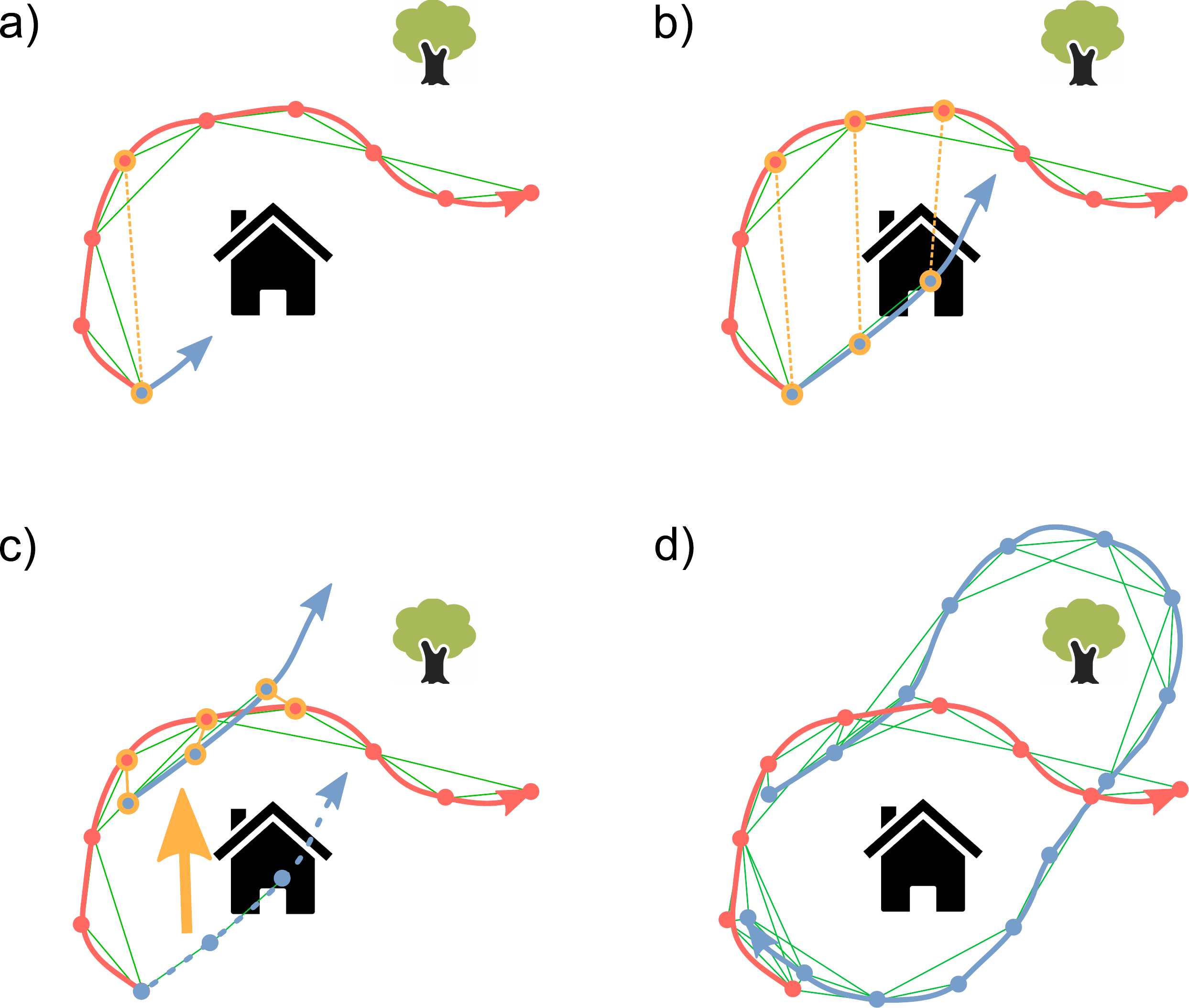}
   \caption{Stages of relocalization in a previous SLAM keyframe map (red). New SLAM map is shown in blue. Yellow circles and lines depict found correspondences of keyframes between both maps.
   a) Initially, both maps are independent. A first keyframe correspondence is found.
   b) Successive relocalization hypothesis building.
   c) After sufficient successive correspondences are found in a hypotheses, the new keyframe map is aligned with the previous map.
    d) Continued SLAM after relocalization in a combined keyframe pose graph.}
    \label{fig:reloc_stages}
\end{figure}

Most SLAM methods in literature focus on a single run of exploration in an environment.
In many practical scenarios, however, such as robotics or augmented reality, it is a desirable property of the SLAM system to allow for relocalization in a previously build map and to seamlessly continue SLAM in novel parts of the environment.
The main challenges in relocalization and SLAM continuation are to robustly detect the relocalization event, to combine the previous and new SLAM results, and to achieve online pose tracking in the global map frame (see Fig.~\ref{fig:reloc_stages}).

Our approach finds an overlapping part of the new trajectory~$\boldsymbol{\xi}^{\mathit{new}}$ with the previous SLAM trajectory~$\boldsymbol{\xi}^{\mathit{prev}}$.
The VI odometry and SLAM methods initially reference the new trajectory in an arbitrary world frame usually set to the identity transform.
On each new keyframe that is included in the new SLAM map, we detect and verify alignments of the keyframe with the old SLAM map using our loop closure detection approach (Sec.~\ref{sec:loopclosuredetection}).
We maintain a set of loop closure hypotheses~$\mathcal{H} = \left\{ h_1, \ldots, h_N \right\}$ which contains hypotheses~$h_n$ representing trajectory parts in the new SLAM map that have been associated with keyframes of the previous map.

If a match is found for a new keyframe~$k^{\mathit{new}}$ with keyframe~$k^{\mathit{prev}}$ in the previous map, we include a new hypothesis $h = \left\{ \left( k^{\mathit{new}}, k^{\mathit{prev}}, \leftidx{_{k^{\mathit{prev}}}}{\boldsymbol{\xi}^\mathit{cls}_{k^{\mathit{new}}}} \right) \right\}$ into~$\mathcal{H}$.
We merge hypothesis by combining their matchings into a single set, if they contain matches of keyframes $\left( k^{\mathit{new}}, k^{\mathit{prev}}\right)$ and $\left( k'^{\mathit{new}}, k'^{\mathit{prev}} \right)$ that are time-sequential in the same order in both the previous and new map, i.e. $\left\| k'^{\mathit{new}} - k^{\mathit{new}} \right\| = 1$ and $k'^{\mathit{new}} - k^{\mathit{new}} = k'^{\mathit{prev}} - k^{\mathit{prev}}$.

As soon as there is a hypothesis with a sufficiently robust number of subsequent matchings (4 in our experiments), we determine the relative pose~$\boldsymbol{\xi}_{\mathit{maps}} = \leftidx{_{k^{\mathit{prev}}}}{\boldsymbol{\xi}^\mathit{cls}_{k^{\mathit{new}}}}$ between the new and previous SLAM map from the most recent keyframe match.
We combine the new SLAM graph (keyframe poses and relative pose constraints) with the previous SLAM graph by transforming all keyframe poses in the new SLAM map to the previous map frame.
Consequently, we optimize the joint objective
\begin{multline}
E_{S}(\boldsymbol{\xi}) =
        E_{S}(\boldsymbol{\xi}^{\mathit{prev}}) + E_{S}(\boldsymbol{\xi}^{\mathit{new}})\\
    + \sum_{\left(k,k'\right) \in \mathcal{C}_{\mathit{maps}}} \mathbf{e}_{\mathit{cls}}^{k,k'}(\boldsymbol{\xi})^\top \mathbf{\Omega}_{\mathit{cls}}^{k,k'} \mathbf{e}_{\mathit{cls}}^{k,k'}(\boldsymbol{\xi}),
\end{multline}
where~$\boldsymbol{\xi}$ concatenates previous and new keyframe poses and~$\mathcal{C}_{\mathit{maps}}$ is the set of loop closure constraints between the previous and new map parts.
After the merging, our loop closure mechanism automatically
creates additional constraints between the map parts which further improves the alignment of the trajectories.

\section{Experiments}

\begin{figure}[t]
  %\includegraphics[width=\linewidth]{eval/experiment-images/ex3_2.png}
  %\hspace{3mm}
  %\includegraphics[width=0.9\linewidth]{setup.pdf} \\    
  \includegraphics[width=0.49\linewidth]{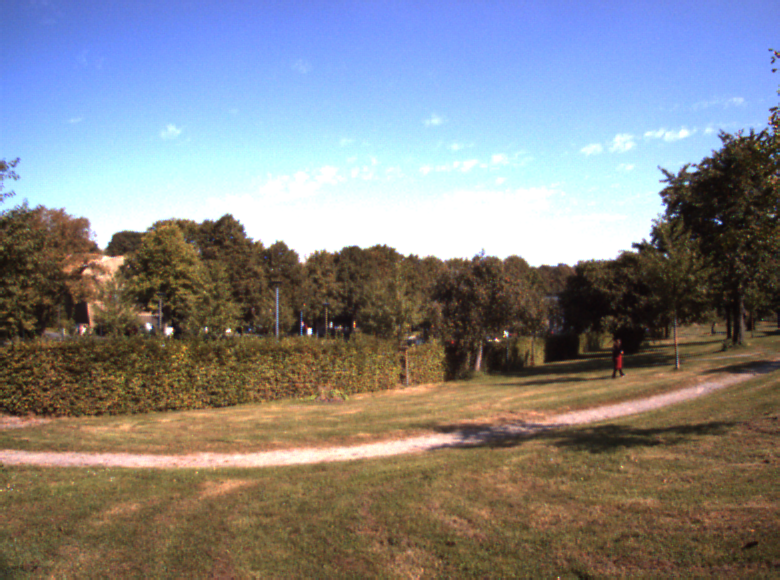}   
  \includegraphics[width=0.49\linewidth]{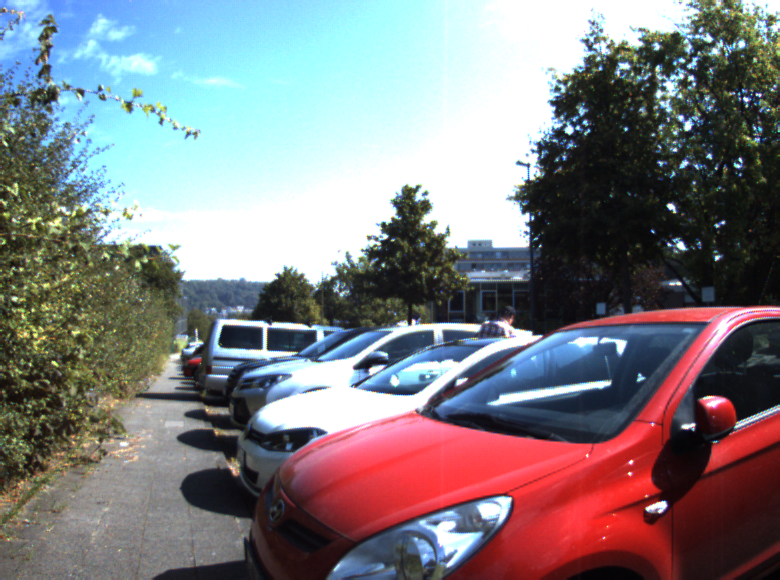}
  \caption{Example images from the outdoor sequences used during the evaluation.}
  \label{fig:sequence_examples}
\end{figure}

We evaluate our visual-inertial SLAM and relocalization approach on the publicly available EuRoC MAV benchmark dataset~\cite{burri2016_eurocmav} and on home-made large-scale outdoor sequences (see Fig.~\ref{fig:sequence_examples} for example images from the dataset).
The EuRoC MAV dataset consists of 11 visual-inertial sequences that have been recorded from a microcopter in two indoor environments, Machine Hall (MH, 5 sequences) and Vicon Room (VR, 6 sequences).
The sequences have been recorded with a visual-inertial sensor unit~\cite{nikolic2014_visensor} that provides time-synchronized 2-axis gyroscope and accelerometer readings at 200\,Hz with WVGA grayscale global shutter stereo camera images at 20\,Hz.
The dataset also includes extrinsic and intrinsic calibration of the sensor as well as ground truth trajectories which have been obtained using a Vicon motion capture system and a Leica laser tracker on the VR and MH sequences, respectively.
Additionally, for the VR sequences, a ground truth 3D laser scanner point cloud is available.

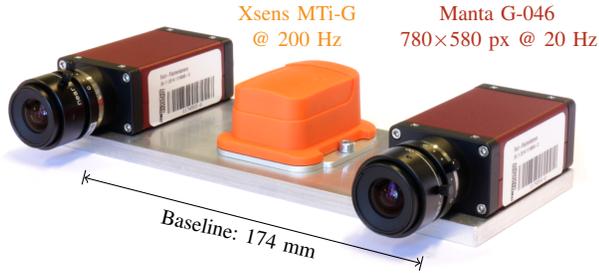
\begin{figure}[t]
  \footnotesize
  \begin{overpic}[trim={0 0 0 0},clip,width=\linewidth,tics=10]{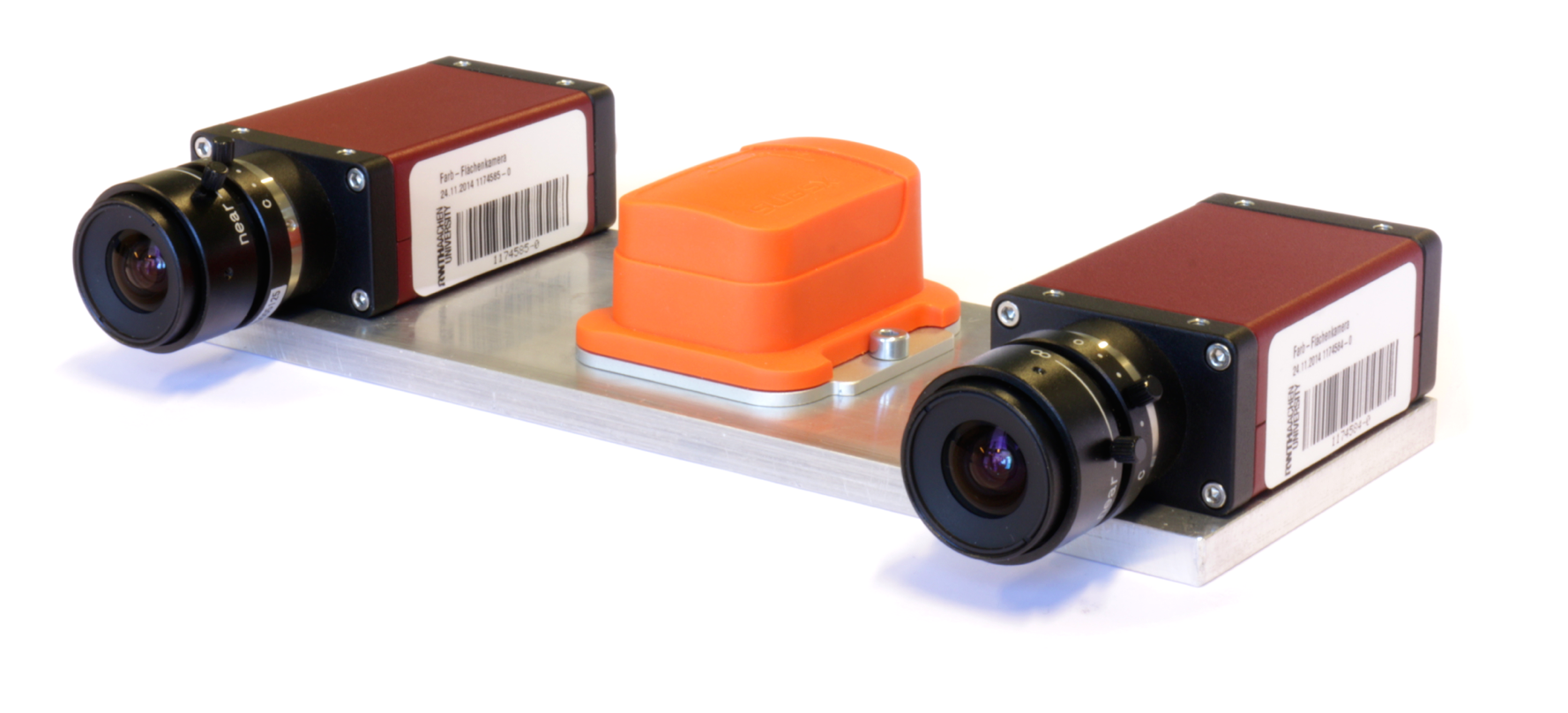}
	\put (40,43) {\textcolor{BurntOrange}{Xsens MTi-G}}
	\put (42,39) {\textcolor{BurntOrange}{@ 200 Hz}}	
	\put (71,43) {\textcolor{BrickRed}{Manta G-046}}
	\put (65,39) {\textcolor{BrickRed}{780$\times$580 px @ 20 Hz}}
	
	\put (28,9) {\rotatebox[origin=c]{-13}{Baseline: 174 mm}}
	
  \begin{tikzpicture}
    \node[anchor=east] at (-0.5,0) {~};
    \tikzstyle{ann} = [fill=white,font=\footnotesize,inner sep=1pt]
    \node[coordinate] (origin) at (0,0) {};
    %\node[] {} [above right=2.5cm and 1cm of origin] (left) {l};
    	\node (left) [above right=1.45cm and 0.5cm of origin] {};
    	\node (right) [above right=0.25cm and 5.2cm of origin] {};
    \draw[arrows=|<->|](left)--(right);

    %\draw[arrows=|<->|](2.6,3)--(7.2,2);
    %\node[ann] at (4,2) {\rotatebox{-12}{Baseline: 174 mm}};
    %\node[inner sep=0pt,rotate=-12,%
    %                  right=0.5cm,minimum height=12pt](f\k) at (d\k) {\t};
  \end{tikzpicture}

	\end{overpic}
  \caption{Visual inertial sensor unit used to record outdoor sequences. Inertial Measurement Unit (IMU) in the middle, surounded by color cameras left and right.}
  \label{fig:setup}
\end{figure}

The second dataset has been recorded using a visual-inertial sensor unit that is built from an Xsens MTi-G IMU providing 3-axis gyroscope and accelerometer data at 200 Hz and two Manta G-046 cameras operating at 20\,Hz and 780$\times$580 resolution in stereo configuration with a baseline of 174 mm.
The setup is depicted in Fig.~\ref{fig:setup}.
Both kinds of sensors provide time-synchronized measurements and are calibrated intrinsically and extrinsically using the Kalibr framework~\cite{furgale2013_kalibr}.
These recordings focus on outdoor environments with challenging lighting conditions and larger scale trajectories than in the EuRoC MAV dataset.
It includes 4 sequences of increasing length and difficulty.
The three sequences (Out1: 875\,m, Out3: 1366\,m, Out4: 1341\,m) have been recorded by a person walking in an urban area.
The person was holding the visual-inertial sensor at a height of $\sim$\,1\,m and pointing the camera in walking direction.
In sequence Out2 (2891\,m), the sensor unit has been mounted on a bicycle steering wheel pointing the camera forward.
In all four sequences, starting and endpoint of the trajectories coincide to enable the verification of the start-to-end trajectory consistency.
Additionally, similar keyframes have been manually marked at checkpoints at which the camera revisits corresponding locations.

\subsection{Simultaneous Localization and Mapping}

We evaluate the performance of our VI SLAM method in relation to the VI odometry approach~\cite{leutenegger2014keyframe} in monocular (only left images) and stereo configuration.
We also compare our stereo method with Stereo LSD-SLAM~\cite{engel2015_stereo_lsdslam} which is a purely vision-based method.
We provide quantitative results in terms of absolute trajectory error (ATE,~\cite{sturm2012_tumrgbdbenchmark}) on the EuRoC MAV benchmark sequences in Table~\ref{tab:slamate}.
In most of the cases, our SLAM approach provides improved ATE.
It can be seen that using stereo images consistently outperforms the monocular setting.
Using the stereo images, the improvement on the VR sequences is not that strong which is due to the small length of the trajectories for which loop closures add less gain in accuracy.
Note that the baseline visual-inertial odometry method itself fails on VR23. 
Due to the rapid motion in this sequence, the images are blurred over many frames, so that interest point detection and matching is less accurate and reliable.
Loop closing and pose graph optimization cannot correct the tracking failure of the underlying visual-inertial method.
Revisiting the interest point detector or using direct methods could alleviate this problem.
Our stereo SLAM system also outperforms the purely vision-based method in~\cite{engel2015_stereo_lsdslam} in two of three cases.
In the V13 sequence, the direct method seems to have an advantage over our keypoint-based method.
It seems that direct image alignment handles motion blur better in this case.

\begin{table}[t]
\centering
\footnotesize
\begin{tabular}{lcc}
\toprule
Sequence & Out3 & Out4           \\
\midrule
VI odometry (stereo) \cite{leutenegger2014keyframe}            & 10.9 m   & 40.7 m         \\
VI SLAM (stereo)             &  \textbf{0.89 m}  & \textbf{0.93 m} \\
\bottomrule
\end{tabular}
\caption{Average checkpoint error in meters.}
\label{tab:checkpoints}
\end{table}

\begin{figure*}[!h]
	\begin{tabular}{c c}
	\begin{overpic}[width=0.45\linewidth,tics=10]{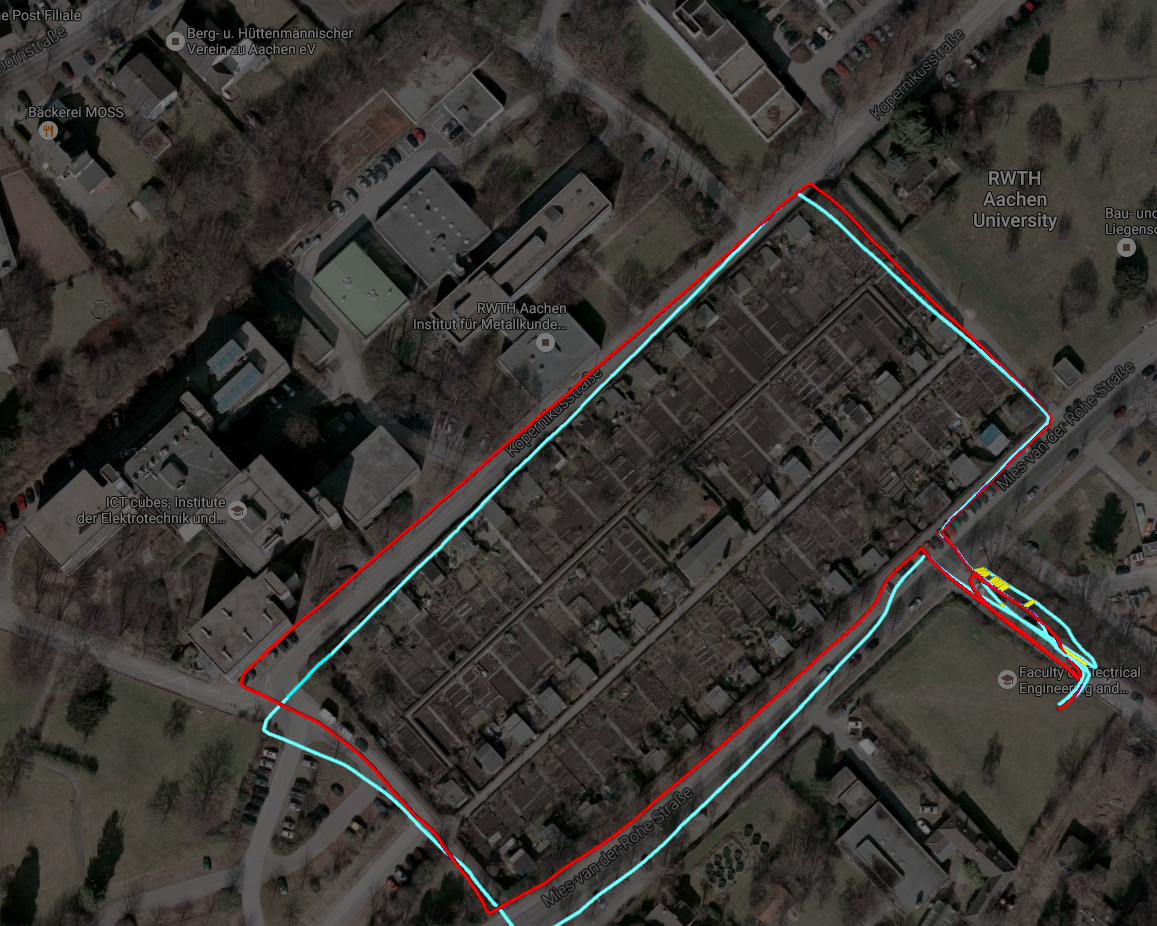}
	\put (5,73) {\textcolor{white}{Sequence Out1}}
	\end{overpic} &
	\begin{overpic}[trim={0 34 0 34}, clip, width=0.45\linewidth,tics=10]{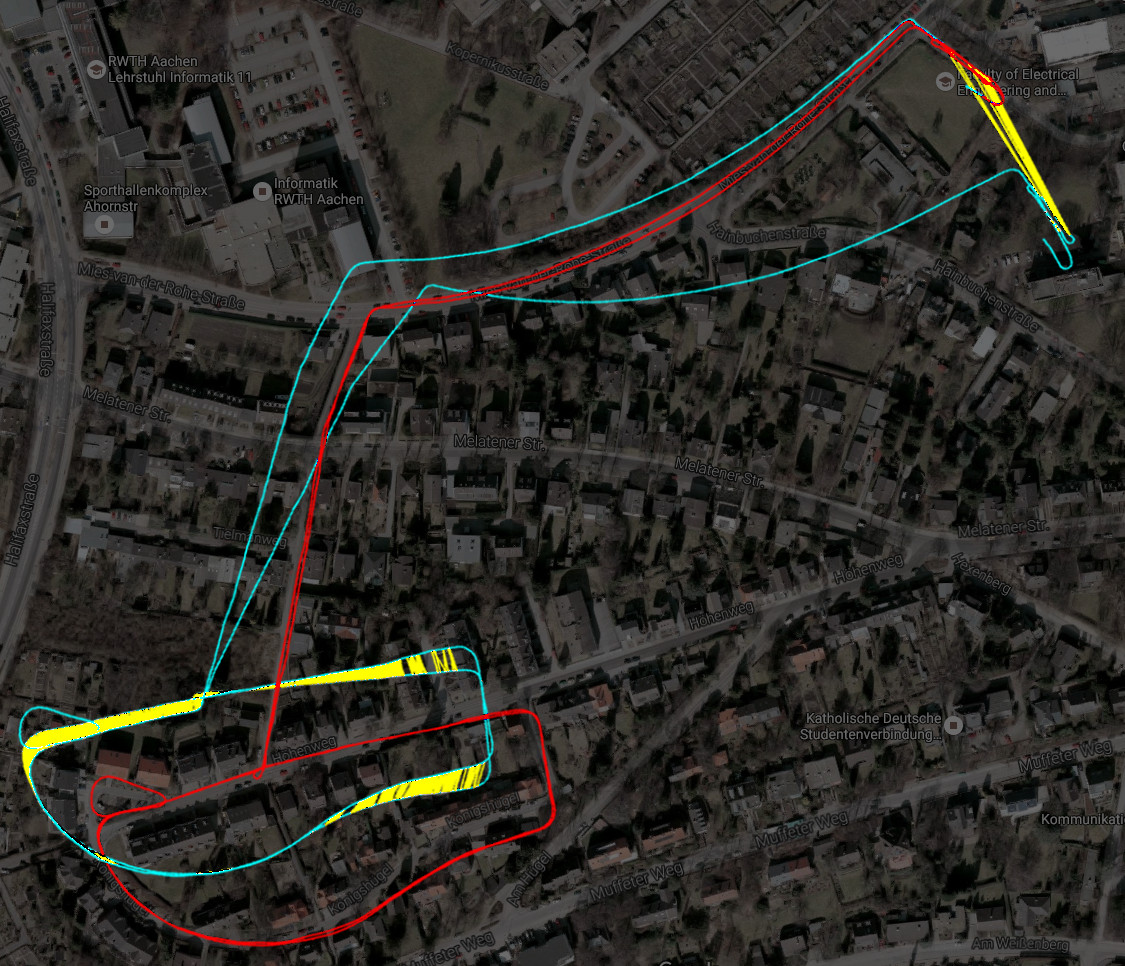}
	\put (5,73) {\textcolor{white}{Sequence Out2}}
	\end{overpic} \\
  	\includegraphics[width=0.47\linewidth]{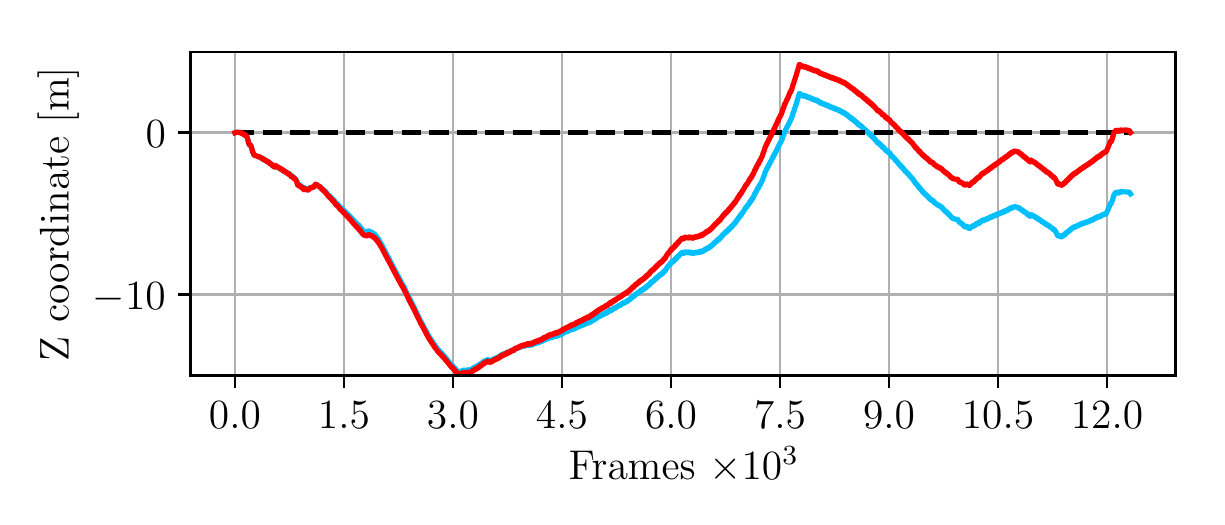}  & \includegraphics[width=0.47\linewidth]{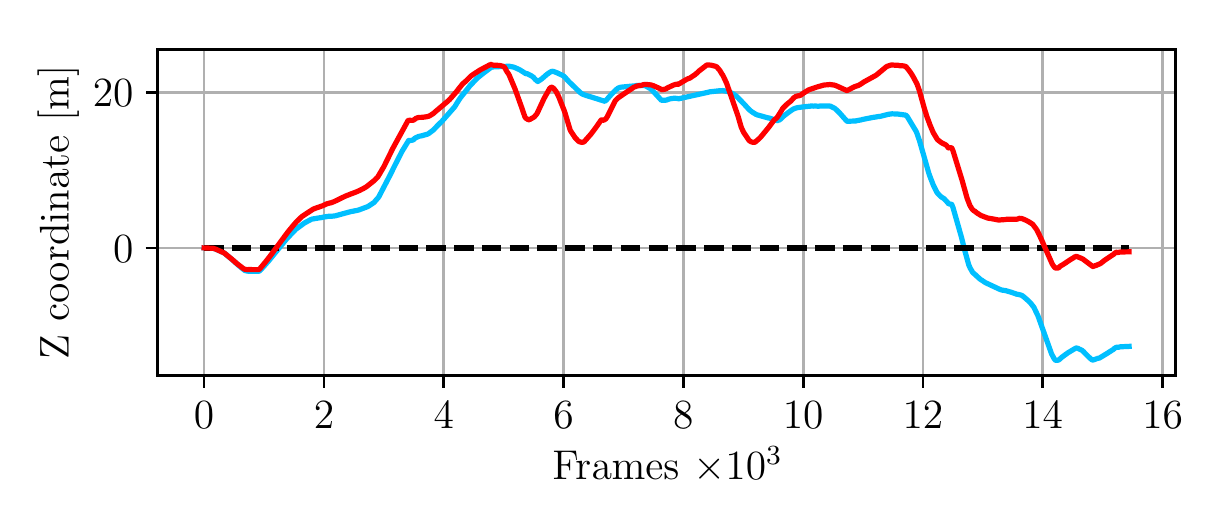} \\
  
  	\begin{overpic}[trim={0 0 0 0}, clip, width=0.45\linewidth,tics=10]{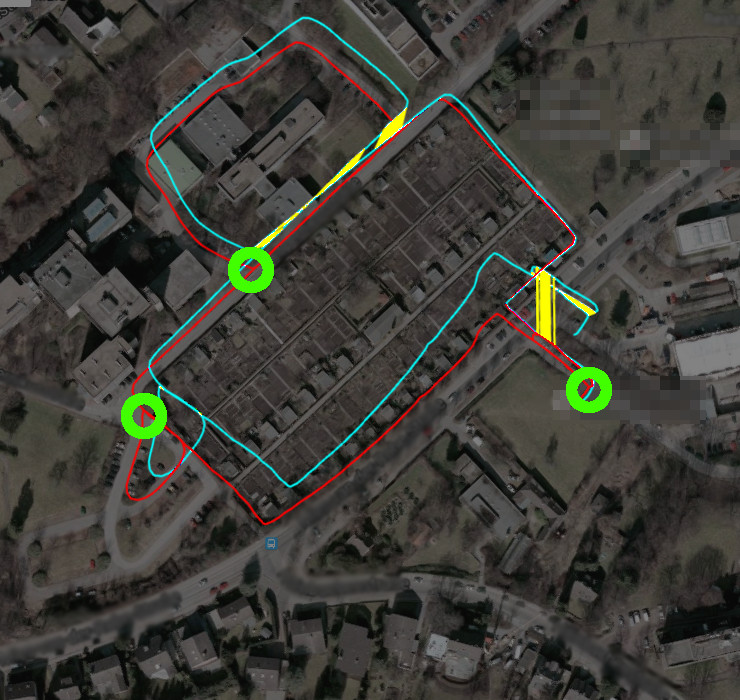}
	\put (5,87) {\textcolor{white}{Sequence Out3}}
	\end{overpic} &
	\begin{overpic}[trim={0 0 0 0}, clip, width=0.45\linewidth,tics=10]{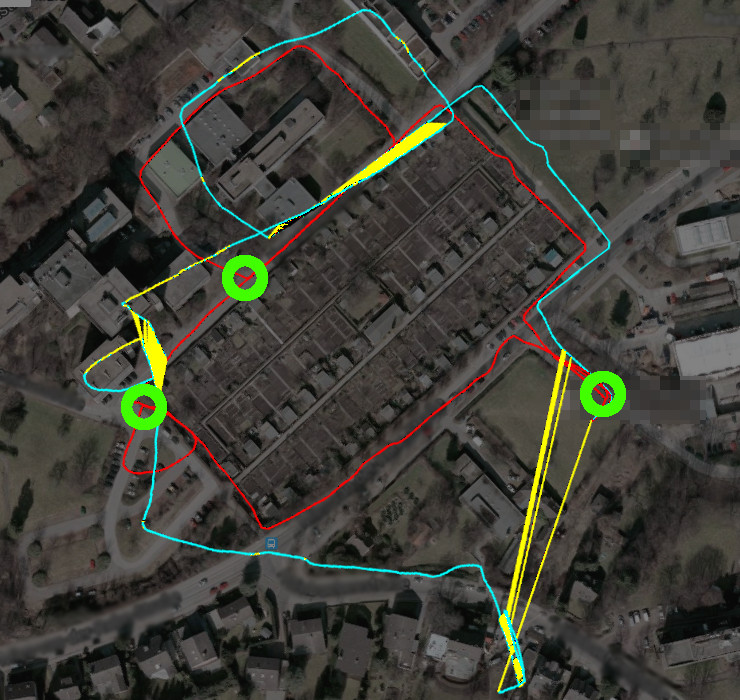}
	\put (5,87) {\textcolor{white}{Sequence Out4}}
	\end{overpic} \\
  	\includegraphics[width=0.47\linewidth]{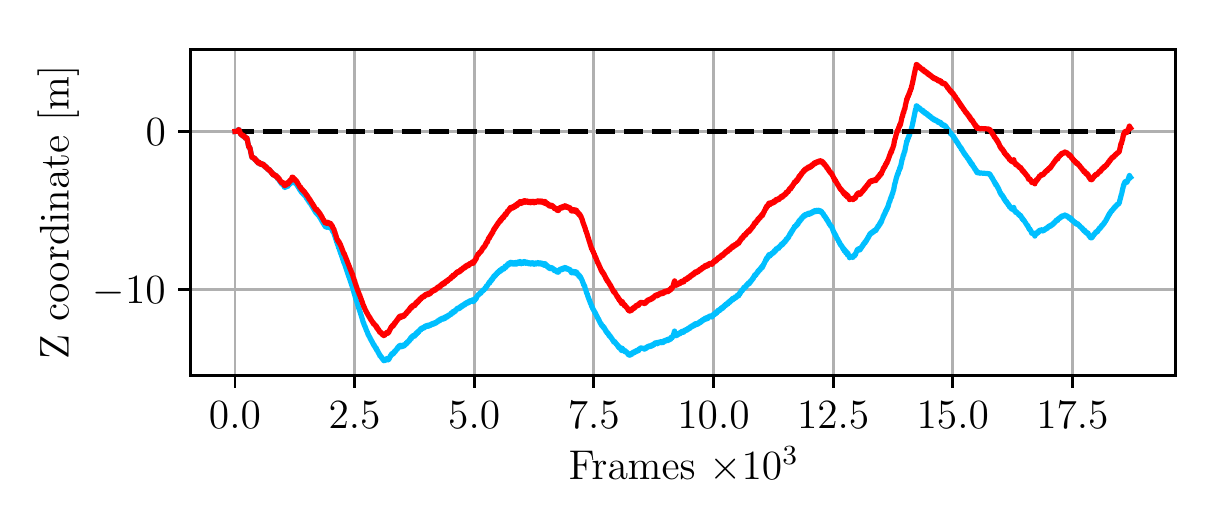}	& \includegraphics[width=0.47\linewidth]{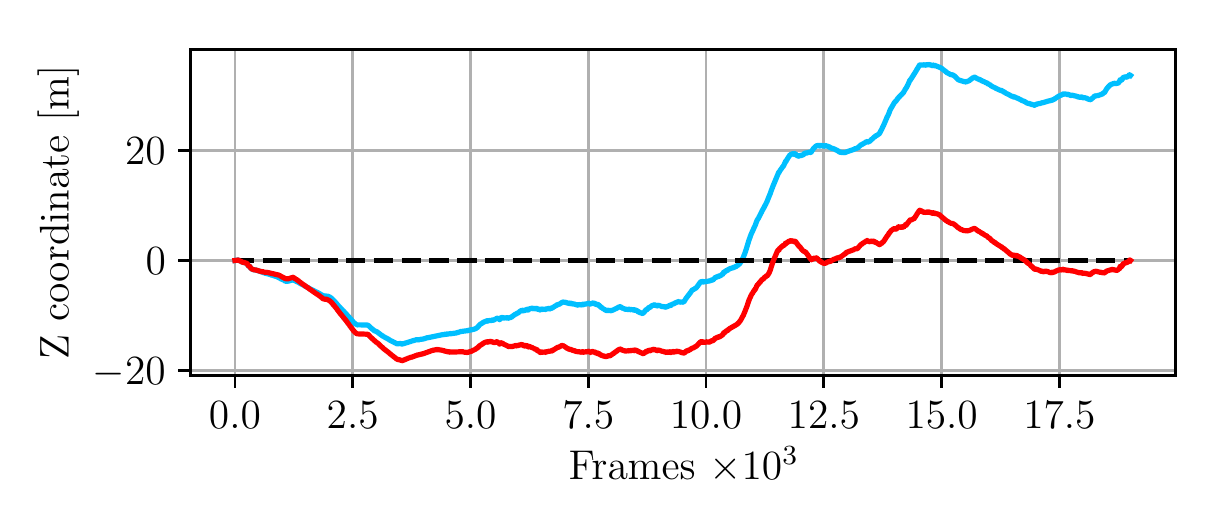} \\
  \label{fig:o1_result}
  \label{fig:o2_result}
  \label{fig:o4_result}
  \end{tabular}
  \caption{VI odometry (blue, \cite{leutenegger2014keyframe}) and SLAM (red) trajectory estimates on our sequences (top-down view and height profile). Yellow lines depict loop closure constraints.}
  \label{fig:o1234_result}
\end{figure*}

Fig.~\ref{fig:o1234_result} shows results of the VI odometry and our SLAM method on the outdoor sequences (Out1-Out4).
Table~\ref{tab:checkpoints} gives quantitative results for the alignment of the trajectories at the checkpoints in Out3 and Out4.
In Out1, both methods recover the medium-scale trajectory loop well, whereas only the SLAM method is able to detect loop closures and aligns starting and endpoint much more accurately.
Sequences Out3 and Out4 contain additional smaller trajectory loops and are approximately double the total length than Out1.
In both sequences it can be seen that our VI SLAM method recovers the shape of the trajectory and the checkpoints and starting/endpoints align well.
In Out4, the VI odometry method exhibits larger drift at several points on the trajectory due to missing camera frames.
Nevertheless, the loop closure constraints in our SLAM method allow our approach to recover from these estimation errors.
Sequence Out2 is a challenging long trajectory recorded on a bicycle.
Here, the VI odometry drifts significantly at a turning point where the orientation of the camera was changed rapidly.
Our SLAM method corrects this drift after the loop closures at the crossing and at the end of the trajectory is detected.

We also tested the monocular visual SLAM methods LSD-SLAM~\cite{engel2014_lsdslam} and ORB-SLAM~\cite{murartal2015_orbslam} using their open-source reference implementations.
Both methods fail to track the camera motion on our sequences.
In the sequences, the camera moves mainly in forward direction, providing less parallax than for instance a sidewards moving camera. 
Tracking often fails when the camera turns quickly to the left or right on the spot.

\subsection{Relocalization}

\begin{table*}[t]
\centering
\footnotesize
\begin{tabular}{lcccccccccc}
\toprule
Sequence & MH1           & MH2           & MH3           & MH4           & MH5           & VR11           & VR12   & VR13 & VR21           & VR22  \\
\midrule
VI odometry \cite{leutenegger2014keyframe}, mono       & 0.34          & 0.36          & 0.30          & 0.48          & 0.47          & 0.12          & 0.16 &    0.24     & \bf 0.12 & 0.22          \\
VI SLAM, mono & \bf 0.25 & \bf 0.18 & \bf 0.21 & \bf 0.30 & \bf 0.35 & \bf 0.11 & \bf 0.13 & \bf 0.20 & \bf 0.12 & \bf 0.20 \\
\midrule
VI odometry \cite{leutenegger2014keyframe}, stereo      & 0.23          & 0.15          & 0.23          & 0.32          & 0.36          & \textbf{0.04} & 0.08 & 0.13 & \textbf{0.10}          & \textbf{0.17} \\
Stereo LSD SLAM~\cite{engel2015_stereo_lsdslam} & n/a & n/a & n/a & n/a & n/a & 0.07 & 0.07 & \bf 0.09 & n/a & n/a  \\
VI SLAM, stereo       & \textbf{0.11} & \textbf{0.09} & \textbf{0.19} & \textbf{0.27} & \textbf{0.23} & \textbf{0.04}   & \textbf{0.05} & 0.11 & \textbf{0.10} & 0.18 \\
\bottomrule
\end{tabular}
\caption{ATE results in meters on the EuRoC MAV benchmark.}
\label{tab:slamate}
\end{table*}

\begin{figure}[t]
  %\vspace{-9px}
  \includegraphics[width=0.495\linewidth]{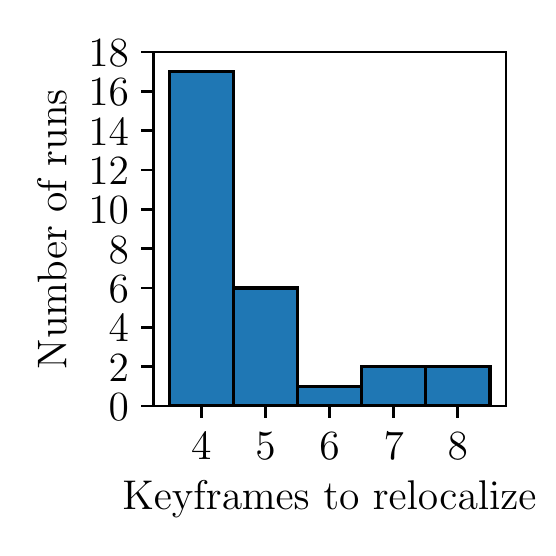}
  \includegraphics[width=0.495\linewidth]{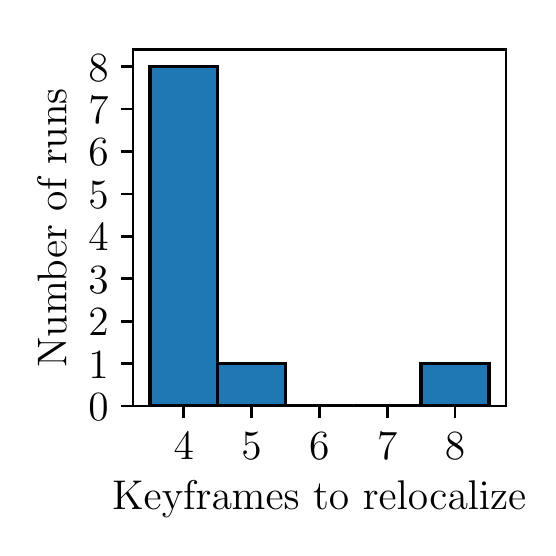}
  \caption{Number of keyframes needed to relocalize within the same sequence (Left: EuRoC MAV, Right: Out1 and Out3).}
  \label{fig:reloc_numkeyframes}
\end{figure}

\begin{figure}[ht]
  \includegraphics[width=0.495\linewidth]{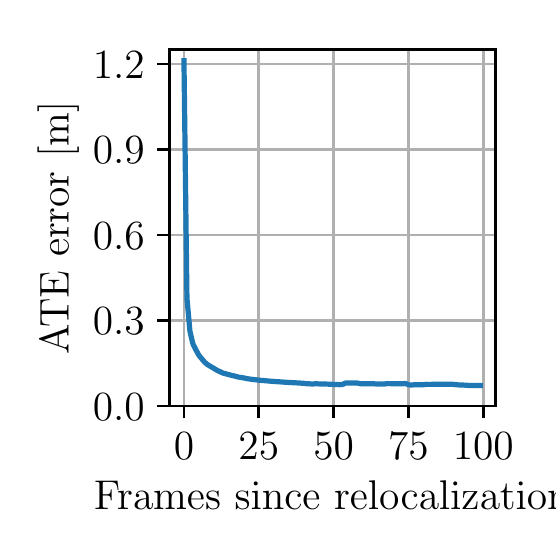}
  \includegraphics[width=0.495\linewidth]{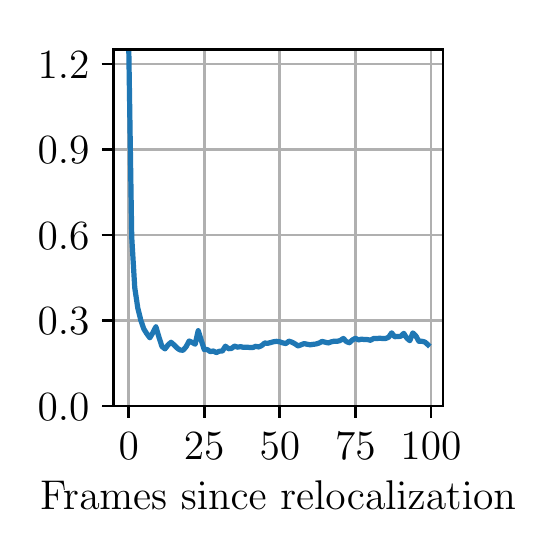}
  \caption{Progression of ATE error in meters after relocalization (Left: EuRoC MAV. Right: Sequence Out1 and Out3).}
  \label{fig:reloc_ateprog}
\end{figure}

\begin{figure}[t]
  \includegraphics[width=1\linewidth]{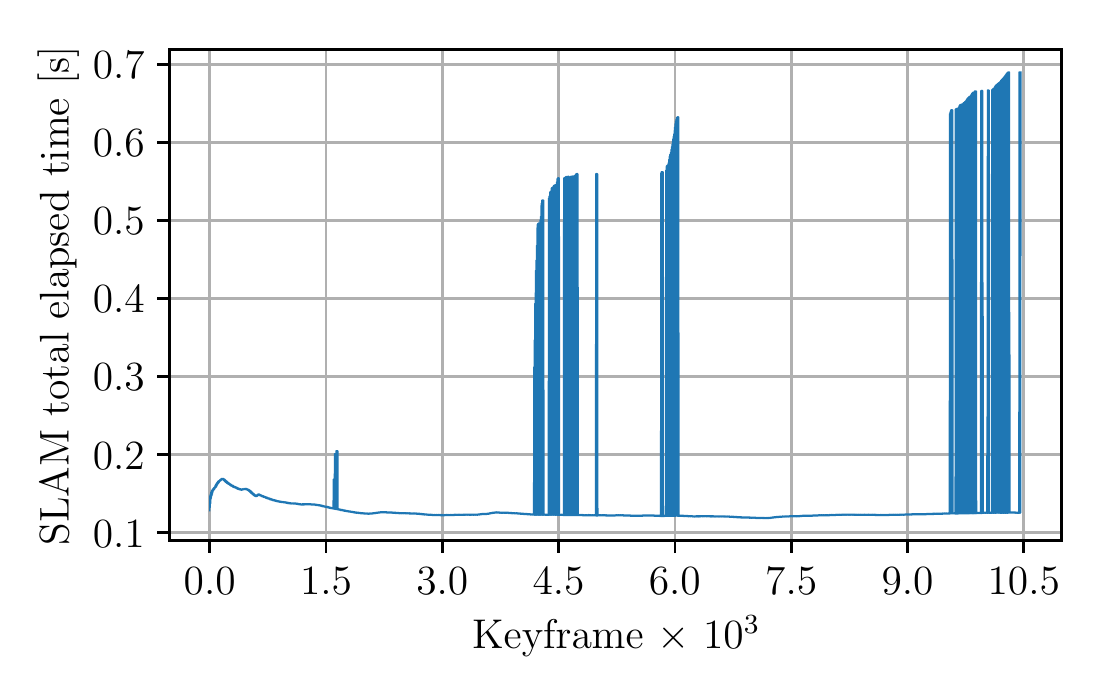}
  \caption{Timing profile of SLAM processing (loop closure detection + pose graph optimization) on the Out3 sequence.}
    \label{fig:runtime_profile}
\end{figure}

\begin{table}[t]
\centering
\footnotesize
\begin{tabular}{lcccc}
\toprule
Previous sequence      & VR12 & VR11    & VR22 & VR21    \\
New sequence      & VR11 & VR12    & VR21 & VR22    \\
\midrule
Combined SLAM & 0.11 m & 0.13 m    & 0.08 m       & 0.13 m \\
\bottomrule
\end{tabular}
\caption{SLAM ATE error in meters after relocalizing a sequence with respect to another sequence.}
\label{tab:reloc_slam}
\end{table}

\begin{table}[t]
\vspace{5mm}
\centering
\footnotesize
\begin{tabular}{lccc}
\toprule
Processing step & VI odometry & Loop closure det. & Graph optim. \\
\midrule
EuRoC MAV   & 6 ms & 87 ms    & 95 ms     \\
Out3      & 9 ms &  116 ms   & 582 ms      \\
\bottomrule
\end{tabular}
\caption{Average run-time for the steps of our method.}
\label{tab:runtime}
\end{table}

We also evaluate the relocalization capability of our SLAM approach.
In a first set of experiment, we relocalize the EuRoC MAV sequence MH1-5 and VR11, VR12, VR21, VR22 and the Out1 and Out3 sequences with respect to themselves in order to test the algorithm in the ideal case.
We start the sequence at several time offsets from its first frame at an interval of 10\,s.
Fig.~\ref{fig:reloc_numkeyframes} gives the histogram of the number of required keyframes to relocalize.
It can be seen that while we use a matching sequence length of 4 to verify a relocalization sequence, our approach relocalizes already within only 9 keyframes from the initialization.
We also determine how the metric relocalization error proceeds after relocalization in terms of ATE towards the ground-truth trajectory (for Out1-Out4 we used the initial SLAM trajectory as ground truth).
Fig.~\ref{fig:reloc_ateprog} shows the results which demonstrate that our method finds an accurate relocalization which settles after only about 20 frames (corresponding to approx. 1\,s).
After the relocalization, our SLAM method continues to include new loop closures of the new trajectory part with the previous map which improves the alignment of the trajectory with the previous map over time.
Note that the results are biased by the SLAM ATE of the first run and should be seen in relation to the achieved SLAM ATE of the sequence.

In a second experiment, we evaluate how well our method aligns a sequence in a previous SLAM map obtained from a different sequence in the VR sequences.
To this end, we measure the SLAM ATE error towards the ground-truth of the combined SLAM map from both sequences.
Table~\ref{tab:reloc_slam} gives the corresponding results.
The resulting ATE is very similar to the ATE achieved by the previous SLAM map alone.

\subsection{Run-Time}

Our VI SLAM approach achieves real-time camera tracking performance on the datasets.
Table~\ref{tab:runtime} lists average run-time measured on the EuRoC MAV sequences and the Out3 sequence.
For profiling, we used a desktop PC with an Intel(R) Core(TM) i7-3770 CPU @ 3.40GHz.
It can be seen that while SLAM is updated at a slower rate in a parallel thread, VI odometry continues to process images at a run-time within the frame-rate.
As can be observed from Fig.~\ref{fig:runtime_profile}, the SLAM run-time peaks to values of up to 0.7\,s when a loop closure is detected and the SLAM pose graph optimization is invoked.
Notably, we run pose graph optimization until convergence, which could also be distributed across the sequence by optimizing only for a few iterations in each frame.

\section{Conclusions}

In this paper, we have proposed a real-time capable visual-inertial SLAM system with the capabilities to relocalize in a previously built map and to continue SLAM in previously unvisited regions.
Our method builds on a state-of-the-art keyframe-based visual-inertial odometry method~\cite{leutenegger2014keyframe}.
We use a state-of-the-art image retrieval method in order to detect potential revisits of places.
Geometric verification and pose estimation is performed using 2D-to-3D RANSAC in order to achieve high precision.
For relocalization, we have proposed a simple but effective multi-frame verification method.
Our SLAM approach allows for seamless combination of previous and new SLAM maps at relocalization and continuation of SLAM afterwards.

In our experiments, our SLAM approach yields state-of-the-art trajectory accuracy compared to the visual-inertial odometry method on an indoor MAV benchmark dataset.
On several challenging outdoor sequences, we demonstrated that our method is able to detect trajectory loops and improve the global consistency of the trajectory estimate.
Our method can quickly relocalize in a previous map using the loop closure detection technique and determines an accurate localization and SLAM estimate afterwards.
Our run-time analysis additionally demonstrates that our method is real-time capable even in long trajectories with several loops.

While we observed high precision in our experiments, our system may detect false positive loop closures due to perceptual aliasing (different locations that appear very similar in corresponding keyframes).
In future work, we want to investigate approaches to reduce the chance of false positives in such cases, for example, using multi-frame context or robust pose graph optimization techniques such as switchable constraints~\cite{suenderhauf2012_switchable}.

\vfill\eject

{\small
\bibliographystyle{IEEEtran}
\bibliography{abbrev_short,egbib}
}

\end{document}